%% file: conference_101719.tex
\documentclass[conference]{IEEEtran}
\IEEEoverridecommandlockouts
\usepackage{cite}
\usepackage{amsmath,amssymb,amsfonts}
\usepackage{algorithmic}
\usepackage{graphicx}
\usepackage{textcomp}
\usepackage{xcolor}
\usepackage{float}
\usepackage{rotating}
\usepackage{ upgreek }
\usepackage{mathtools}

\usepackage{graphicx}
\usepackage[font={sf,scriptsize},           
            labelfont={bf,color=accessblue},
            caption=false]                  
           {subfig}

\usepackage{comment}

\newsavebox\mybox

\def\BibTeX{{\rm B\kern-.05em{\sc i\kern-.025em b}\kern-.08em
    T\kern-.1667em\lower.7ex\hbox{E}\kern-.125emX}}
\begin{document}

\title{
Transfer Learning Based Efficient Traffic Prediction with Limited Training Data\\
}

\author{\IEEEauthorblockN{Sajal Saha, Anwar Haque, and\ *Greg Sidebottom}
\IEEEauthorblockA{\textit{Department of Computer Science} \\
\textit{University of Western Ontario, London, ON, Canada}\\
\textit{*Juniper Networks, Kanata, ON, Canada}\\
Email:\{ssaha59, ahaque32\}@uwo.ca, *gsidebot@juniper.net}
}

\maketitle

\begin{abstract}
Efficient prediction of internet traffic is an essential part of Self Organizing Network (SON) for ensuring proactive management. There are many existing solutions for internet traffic prediction with higher accuracy using deep learning. But designing individual predictive models for each service provider in the network is challenging due to data heterogeneity, scarcity, and abnormality. Moreover, the performance of the deep sequence model in network traffic prediction with limited training data has not been studied extensively in the current works. In this paper, we investigated and evaluated the performance of the deep transfer learning technique in traffic prediction with inadequate historical data leveraging the knowledge of our pre-trained model. First, we used a comparatively larger real-world traffic dataset for source domain prediction based on five different deep sequence models: Recurrent Neural Network (RNN), Long Short-Term Memory (LSTM), LSTM Encoder-Decoder (LSTM\_En\_De), LSTM\_En\_De with Attention layer (LSTM\_En\_De\_Atn), and Gated Recurrent Unit (GRU). Then, two best-performing models, LSTM\_En\_De and LSTM\_En\_De\_Atn, from the source domain with an accuracy of 96.06\% and 96.05\% are considered for the target domain prediction. Finally, four smaller traffic datasets collected for four particular sources and destination pairs are used in the target domain to compare the performance of the standard learning and transfer learning in terms of accuracy and execution time. According to our experimental result, transfer learning helps to reduce the execution time for most cases, while the model's accuracy is improved in transfer learning with a larger training session.

\end{abstract}


\section{Introduction}
Over the past decade, internet traffic growth has been accelerated by the rapid and widespread development of network technologies such as the Internet of Things (IoT), Industrial IoT (IIoT), 5G, and cloud computing. Also, according to the Cisco Annual Internet Report \cite{12}, there is an indication of a massive number of internet users by 2023 (approximately 5.3 billion). With the advancement of network and communication technologies and the increment of potential users, it is inevitable to design a more intelligent and self-organizing network (SON) capable of adapting the dynamic behavior and taking preemptive actions. 
The ISP (Internet Service Provider) network traffic is a critical metric for assessing network load and performance, and achieving a precise traffic prediction can be a suitable technique for proactive network management \cite{13}. Most network management operations, such as resource allocation, short-term traffic scheduling or re-routing, long-term capacity planning, network architecture, and network anomaly detection, need an accurate traffic prediction tool. However, real-world internet traffic prediction is a challenging area. The main reasons are as follows \cite{6}: 

\subsubsection{Data heterogeneity}
Generally, internet traffic is heterogeneous in the temporal dimension, and a different graphical region can present different traffic patterns at different time-period. Therefore, it is challenging to propose a generic prediction model that can predict other traffic datasets with equal accuracy. 

\subsubsection{Anomalous data}
Real-world internet traffic is sensitive to various external and internal factors that produce non-stationary complex traffic patterns. Also, those events can exhibit data points that are not in the range of the data distribution, hindering the model generalization capability during the deployment. So it is crucial to handle the outliers or anomalous data points before using them to train the prediction model. 

\subsubsection{Data scarcity}
For better prediction on the testing set, it is necessary to train the prediction model with a massive dataset so that model can generalize the pattern well. But in reality, it is challenging to manage a large historical dataset to train the prediction model, which leads to poor performance.  

Predicting internet traffic is commonly thought of as a time series forecasting problem that can be tackled by using traditional statistical techniques such as ARIMA \cite{14}, SARIMA, Holt-Winter \cite{15}, etc. But these models cannot predict the non-linear component of the actual internet traffic. With the advancement of artificial intelligence, machine learning\cite{isncc} and deep learning\cite{icc} based traffic prediction models achieve superior performance improvement over the classical approaches. However, modern machine learning and deep learning algorithms require a massive amount of historical data to learn the general pattern in the traffic. Unfortunately, it is quite difficult to ensure enough data for building a new model separately for different geographical or network sectors. Also, it is unrealistic to design and deploy an individual model for each network segment for a larger ISP as the process is expensive and time-consuming. 

Transfer learning \cite{10} comes into play to solve the problem stated earlier by using the knowledge from the existing prediction model to devise a new model for another dataset. This is the key technology for the industrialization of the deep learning solution. The transfer learning approach consists of three methods: parameter transfer, domain adaption, and multi-task learning. The parameter transfer technique uses the learned parameters from the source data and uses this knowledge to build a model for the target data. This technique can help in two ways: re-use knowledge for quick learning and model smaller datasets. The main contributions of this work are:

\begin{itemize}
 \item Analyzing different deep sequence models in the source domain to identify the best performing model for the target domain.  
\item Investigating transfer learning technique on the target domain consisting of multiple datasets based on the best two models from the source domain. 
\item Comparing the performance of standard learning and transfer learning in terms of average accuracy and execution time in the target domain for comparatively smaller datasets.
 
\end{itemize}

This paper is organized as follows. Section \ref{Literature Review} describes the literature review of current traffic prediction using machine learning models. Section  \ref{Methodology} presents the methodology, including dataset description, deep learning models explanation, anomaly identification process, and experiment details. Section  \ref{Results and Discussion} summarizes the different deep learning methods' performance and draws a comparative picture among prediction models with and without outliers in the dataset. Finally, section  \ref{Conclusion} concludes our paper and sheds light on future research directions.

\section{Literature Review}
\label{Literature Review}
Wu, Qiong, et al. \cite{1} proposed a novel mobile traffic prediction framework that combines the parameter-transfer \cite{7} and domain adaption \cite{8} approaches from deep transfer learning to enhance the model performance with a smaller dataset. The framework functionality is divided into two main parts: build the target prediction model with a massive dataset and then use the pre-trained model knowledge from the source domain, which faces the data-scarcity problem. Furthermore, they applied a GAN-based approach to solving the domain shift problem due to different data distribution between source and target domain. According to their experiment, the GAN-based domain adaption helps their model leverage the knowledge from the source domain to the target domain, giving a better prediction for a smaller dataset. However, the quality of the generated data using GAN is not presented. Moreover, since GAN is suffering to reach a stable training point \cite{9}, it is unclear how their GAN performs in generating samples.

Li, Ning, et al. \cite{2} proposed a satellite traffic prediction model based on Gated Recurrent Unit (GRU) architecture that uses the transfer learning and particle filter algorithm for better prediction with a smaller dataset and lower training time. According to their experimental environment, they used similar distribution for the source and target domain in transfer learning. It is unclear how their prediction model will perform if the source and target distribution are asymmetric. The distribution is unlikely to be similar for the source and target domain in the real world, and that's why it is crucial to validate the model performance with data coming from a different distribution than that source domain. A wireless cellular traffic prediction model has been proposed by Zeng, Qingtian, et al. \cite{3}, which is trained based on a cross-domain dataset. They also used the already trained model's parameters for the target domain by adjusting the parameter's values or transferring the learned features to improve the model accuracy. The experimental results showed the outperformance of the model with the transfer learning capability than the model having no transfer learning. In \cite{4}, Dridi, Aicha, et al. proposed a transfer-learning based deep learning model for time series classification and prediction. The transfer-learning technique is adapted in their model mainly for two reasons: better prediction with a smaller dataset and re-adaption of the already trained model for another domain. Their experimental results showed an outperformance of transfer learning in time-series prediction. However, their source and target domain data are drawn from the same distribution, which is unlikely to happen in the real world. 

Huang, Yunjie, et al. \cite{5} proposed a spatial-temporal traffic prediction model has been proposed with transfer learning capability to predict urban traffic from small to medium-sized cities. They applied graph neural networks to model the traffic network's spatial information, while the temporal information was modeled using a deep sequence model called GRU. Two different datasets have been used as a source and a target domain in the experiment. The experimental results showed the advantage of transfer learning in predicting a smaller dataset. In \cite{6}, the authors proposed an intelligent cellular traffic prediction model based on a graph convolutional network with an attention mechanism. The temporal component of the traffic data has been captured using a convolutional neural network (CNN). They also used the power of transfer learning for better prediction on a large scale. Their experimental results showed the effectiveness of transfer learning in reducing the training cost and re-using the knowledge from the already trained model.  

The current works show the usefulness of the transfer learning method in traffic prediction. But we found a lack of investigation in the performance comparison of the deep sequence model such as RNN and its varieties in real-world traffic prediction based on deep transfer learning. In this work, a comprehensive analysis of different deep sequence models has been performed for networks with limited training data based on transfer learning. In addition, we evaluate the performance using five different real-world internet traffic datasets.

\begin{figure}
    \centering
    \includegraphics[width=9cm, height=9cm]{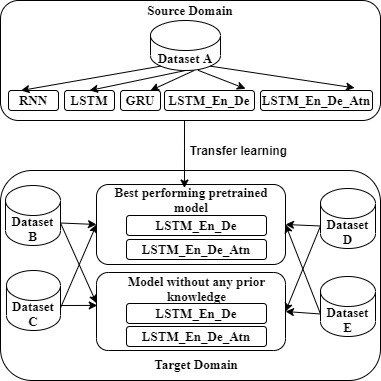}
    \caption{High level presentation of methodology}
    \label{fig:my_label}
\end{figure}

\section{Methodology}
\label{Methodology}
In this section, we first introduce the real IP traffic dataset and the proprocessing steps used in our experiment in subsection \ref{data_preprocessing}. Then, we describe deep transfer learning technique in subsection \ref{deep tl}. After that, we explain deep learning model background in subsection \ref{model}. The model performance evaluation metrics are described in subsection \ref{metric}. Finally, we summarize the configuration of our experimental environment in subsection \ref{software}.

\subsection{Dataset and Preprocessing Steps}
\label{data_preprocessing}
Real internet traffic telemetry on several high-speed interfaces has been used for this experiment. The data are collected every five minutes for a recent thirty days time period. A total of five different datasets are used in our experiment. The source domain dataset, Dataset A, consists of 8563 data samples and it is used to build the predictive model for the source task. The other four datasets, Dataset B, C, D, and E are comparatively smaller in size, having 363, 369, 358, and 365 data instances, respectively. We used these four datasets for the predictive task in target domain.

\subsubsection{Data Windowing}
\begin{figure}[!htbp]
\centering
    \includegraphics[width=9cm,height = 2.5cm]{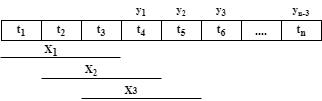}
    \caption{Data windowing}
    \label{fig:Data Windowing}
\end{figure}

Time series data need to be expressed in the proper format for supervised learning. Generally, the time-series data consists of several tuples (time, value), which is inappropriate for feeding them into the machine learning model. So, we restructured our original time series data using the sliding window technique. The sliding window technique is illustrated in Fig. \ref{fig:Data Windowing}. For example, three historical points ($X$) called features are used to predict the next data point ($y$) known as a target in the given sliding window example.

\subsection{Deep Transfer Learning}
\label{deep tl}
Transfer learning is an deep learning or machine learning optimization approach in which knowledge is transferred from one domain to another similar domain. We can formally define transfer learning in terms in terms of domain and task \cite{11}. There are two domains such as source and target domain involved in transfer learning while the domain consists of a feature space $X$ and probability distribution $P(X)$ where $X=\{x_1, x_2,..., x_n\}\in X$. The task $T$ in transfer learning is consists of label space $Y$ and an objective function $f:X \longrightarrow Y$ while $X$ is a feature space for particular domain, $\{X, P(X)\}$. In transfer learning, the source domain $D_S$ and target domain $D_T$ consists of two different task $T_S$ and $T_T$ and the purpose of the transfer learning is to assist the task in target domain to perform better using the knowledge in $D_S$ and $T_S$. 

\begin{figure*}[!htbp]
\centering
    \includegraphics[width=18cm,height = 6cm]{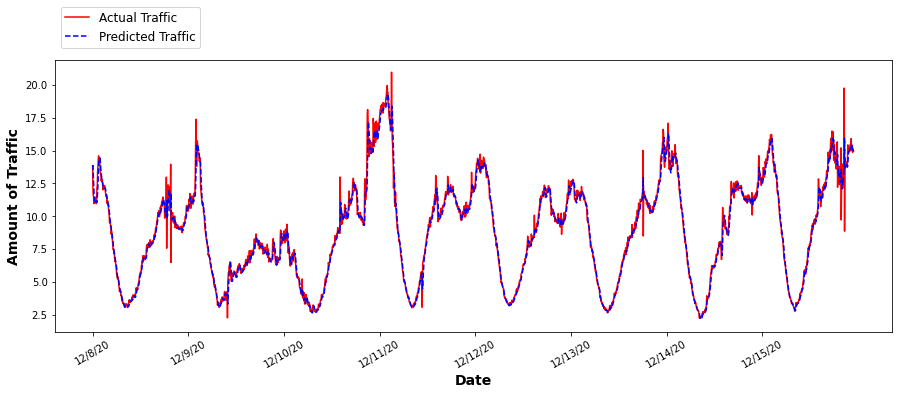}
    \caption{Actual traffic vs predicted traffic in source domain using LSTM\_En\_De model}
    \label{fig:source domain}
\end{figure*}

\subsection{Predictive Models}
\label{model}
\subsubsection{Recurrent Neural Network (RNN)}
At each step, the RNN produces the current output using the current sequence information and the prior sequence information. In the final stage, the model learns knowledge about all previous data points in the series. Text mining, audio classification, language modelling, time series analysis, and other sequential data applications benefit from the RNN paradigm. 


\subsubsection{Gated Recurrent Unit (GRU)} The GRU has been proposed to tackle the RNN model's short-time memory problem. In this architecture, the gate idea is employed to govern the flow of information between two adjacent cells. The update gate in the GRU model determines whether or not the previous cell output should be transferred to the next cell. A gate is a mathematical unit that can decide the relevance of data and whether it should be saved or not. The GRU model has two gates that function on updating the cell state: update gate and reset gate.

\subsubsection{Long Short-Term Memory (LSTM)} The LSTM serves a similar purpose to the GRU model. Along with the update and reset gates, there are two more gates named forget and output gate. The LSTM has more control over information flow between network cells. The LSTM network is widely used for classifying and predicting time-series data. It solves the classic RNN model's vanishing gradient problem and outperforms it in terms of performance.

\subsubsection{LSTM Encoder-Decoder (LSTM\_En\_De)} The sequence-to-sequence model is a model that can predict an output sequence from an input sequence. It is made up of two recurrent neural networks, one of which is an encoder and the other a decoder. The encoder turns the input sequence into a fixed-length context vector, which the decoder receives. The decoder takes the context vector and the encoder's end state as input and produces a series of outputs.

\subsubsection{LSTM Encoder-Decoder with Attention Layer (LSTM\_EN\_DE\_Atn)} This model, also known as the sequence-to-sequence model, may predict an output sequence given an input sequence. However, the encoder-decoder model cannot identify significant contextual relationships from extended sequence data, lowering model performance and accuracy. The encoder-decoder model's extra attention layer\cite{16}, on the other hand, can detect relevance in the long sequence data.

\subsection{Evaluation Metrics}
\label{metric}
We used Mean Absolute Percentage Error (MAPE) to estimate the performance of our traffic forecasting models. The performance metric identifies the deviation of the predicted result from the original data. For example, MAPE error represents the average percentage of fluctuation between the actual value and predicted value. Therefore, we can define our performance metric mathematically as follow: 
\begin{equation}
    MAPE = \dfrac{1}{n}\sum_{i=1}^n \bigg| \dfrac{p_i-o_i}{o_i} \bigg| \times 100 \%
\end{equation}
\begin{equation}
    Accuracy = (100-MAPE)\% 
\end{equation}
Here, $p_i$ and $o_i$ are predicted and original value respectively and  $n$ is the total number of test instance.

\subsection{Software and Hardware Preliminaries}
\label{software}
We used Python and deep learning library TensorFlow-Keras\cite{17} to conduct the experiments.  Our computer has the configuration of Intel (R) i5-9500T CPU@2.20GHz, 8GB memory, and a 64-bit Windows operating system. 

\section{Results and Discussion}
\label{Results and Discussion}
\input{result}

\section{Conclusion}
\label{Conclusion}
In this work, we evaluate the effectiveness of the deep transfer learning technique in real-world internet traffic prediction for the network with smaller training data. It is practically challenging to arrange a large dataset for efficient model training. Furthermore, designing an individual model for each particular host in the large network is infeasible due to time and resource complexity. Therefore, a deep transfer learning based traffic prediction methodology is proposed that is expected to provide better performance with a comparatively smaller dataset. Our experiment used one large dataset for the source domain and four smaller datasets for the target domain. A total of five different deep learning models such as RNN, LSTM, LSTM\_En\_De, LSTM\_En\_De\_Atn, and GRU were implemented to analyze our source domain traffic. Then we transferred knowledge from the best two prediction models (LSTM\_En\_De and LSTM\_En\_De\_Atn) in the source domain to enhance the prediction task’s performance in the target domain. As a result, the target domain learning became significantly faster with the transfer learning approach than the standard learning for most of the target domain’s datasets. Also, transfer learning improves the model accuracy when it is trained for the larger epoch. According to our experiment, deep transfer learning is an efficient approach to predict the non-stationary and non-linear real-world internet traffic even with a smaller size by re-using the knowledge. In the future, we would like to extend this work by adding the multi-step prediction to evaluate transfer learning performance for both single-step and multi-step forecasting.
\bibliographystyle{IEEEtran}
\bibliography{ref}

\vspace{12pt}

\end{document}

%% file: result.tex
\subsection{Experimental setup}
\label{sss:experimental setup}
As we mentioned earlier, our main research objectives are to find out the effectiveness of deep transfer learning in real-world traffic prediction with limited data and high volatility. To conduct this experiment, we used five different internet traffic datasets extracted from real-world telemetry. But all the datasets are not equivalent according to the data volume; four of them have a smaller number of data instances compared to the other. We used dataset $A$ to build our source traffic prediction model to transfer knowledge to the target domain. The other datasets from $B$ to $E$ are considered for the target domain as they have a smaller number of historical data for learning. So, our experiment has two main parts: i.) design source prediction model on dataset $A$ ii) transfer the source domain knowledge to design predictive models for target domains based on dataset $A$, dataset $B$, dataset $C$, and dataset $D$ and compare the performance with the standard learning to determine the effectiveness of deep transfer learning in real-world traffic prediction. 

\input{tables/source_domain}
\input{tables/lstm_en_de}
\input{tables/lstm_en_de_atn}

\subsection{Result analysis}
\label{sss:result}
\subsubsection{Source domain prediction model}
We applied several deep sequence models such as RNN and their variants in this phase, including LSTM, LSTM En\_De, LSTM En\_De\_Atn, and GRU, to identify the best performing models for our source domain. All our model training continued for 100 epochs with a batch size of 16. The prediction accuracy for all the prediction models on dataset $A$ is summarized in Table \ref{tab:source domain}. According to our experimental results, RNN is the worst performing model with an accuracy of 92.49\%, which is improved by more than 2\% and 1\% after applying LSTM and GRU, respectively. Since LSTM and GRU can retain the information from the longer sequence, it is expected to have a better performance than RNN for final prediction. Finally, we achieved our best prediction result using LSTM\_En\_De with an accuracy of 96.06\%, which is around a 4\% improvement compared to the RNN model. The LSTM\_En\_De\_Atn also gave us around 4\% higher accuracy (96.0\%) when compared with the RNN model, which is very close to the LSTM En\_De model performance. Fig. \ref{fig:source domain} depicts the actual traffic vs. predicted traffic by the best-performing model in the source domain. According to our experiment, the encoder-decoder architecture-based models performed better, and we considered our two best-performing models in the second step of our investigation. We used two source domain model-based LSTM En\_De and LSTM En\_De\_Atn to compare and validate the performance of deep transfer learning.

\subsubsection{Target domain prediction model}
In this phase, we design a predictive model for datasets $B$, $C$, $D$, and $E$ using both standard learning and transfer learning approaches. For both strategies, 70\% of the data has been used for the training, while the remaining 30\% has been applied to test the model. Two best-performing models, such as LSTM En\_De and LSTM En\_De\_Atn from the source domain, are used to design the target domain's predictive models. Also, we trained the target domain models for different epoch lengths to identify the best training settings and make a comparative analysis between standard learning and transfer learning in terms of total training time and accuracy. In the case of transfer learning, we freeze the reused layers during the first ten epochs with a more significant learning rate of 0.001, allowing the new layers to learn reasonable weights. Then we unfreeze the reused layers and decrease the learning rate to 0.0001 to avoid damaging the reused weights. Finally, to calculate the average training time and average accuracy of the prediction in the target domain, we executed the experiment five times and took the corresponding metric's average.
\input{tables/lstm_en_de_fig}
The details results for four different data sets are presented in Table \ref{tab:taget_domain_en_de} and Table \ref{tab:taget_domain_en_de_atn} , respectively, for the LSTM En\_De and LSTM En\_De\_Atn model. The average accuracy (Avg. Acc.) and average total training time in second (Time (S)) indicates the mean accuracy and mean training time for five different runs. Fig. \ref{fig:lstm en de model accuracy} and Fig. \ref{fig:lstm en de model time} depicted the comparative analysis of the average accuracy and average training time between standard learning and transfer learning for LSTM\_En\_De model. We presented the comparative analysis of the average accuracy and average training time between standard learning and transfer learning for LSTM\_En\_De\_Atn model in  Fig. \ref{fig:lstm en de atn model accuracy} and Fig. \ref{fig:lstm en de atn model time}. For the LSTM En\_De model, the model training time is much lesser in transfer learning than standard learning for all target domain's datasets. The training time gap between the two learning strategies is increased with the epoch size, which indicates that transfer learning has the chance of quick learning in case of longer training. Also, the graphs indicate better accuracy with transfer learning for the first three epochs settings, such as 250, 200, and 150, for all four datasets. Our results indicate a similar pattern in the LSTM En\_De\_Atn model, although the gap between average time is smaller than in the LSTM En\_De model. Transfer learning gave us better training time or better prediction accuracy for all settings and performed better for more significant epochs in both metrics.
\input{tables/lstm_en_de_atn_fig}

%% file: tables/source_domain.tex
\begin{table}
\centering
\caption{Performance summary for all model in source domain}
\label{tab:source domain}
\label{tab:Performance summary for all model}
\begin{tabular}{||l||c||} 
\hline
\textbf{Model}                                     & \textbf{Accuracy (\%)}  \\ 
\hline
RNN                                       & 92.49          \\ 
\hline
LSTM                                      & 94.97          \\ 
\hline
GRU                                       & 93.51~         \\ 
\hline
LSTM\_En\_De                      & 96.06          \\ 
\hline
LSTM\_En\_De\_Atn & 96.05          \\
\hline
\end{tabular}
\end{table}

%% file: tables/lstm_en_de.tex
\begin{table*}
\caption{Results summary in target domain using LSTM Encoder Decoder Model}
\label{tab:taget_domain_en_de}
\label{table:lstm_en_de}
\centering
\begin{tabular}{|l|l|l|l|l||l|l|l|l||l|l|l|l||l|l|l|l|} 
\hline
\multicolumn{1}{|c|}{} & \multicolumn{4}{c||}{Dataset B}                                                                                                                                                                                                              & \multicolumn{4}{c||}{Dataset C}                                                                                                                                                                                                            & \multicolumn{4}{c||}{Dataset D}                                                                                                                                                                                                               & \multicolumn{4}{c|}{Dataset E}                                                                                                                                                                                                               \\ 
\hline
                       & \multicolumn{2}{l|}{\begin{tabular}[c]{@{}l@{}}Standard \\Learning\end{tabular}}                                     & \multicolumn{2}{l||}{\begin{tabular}[c]{@{}l@{}}Transfer\\Learning\end{tabular}}                                      & \multicolumn{2}{l|}{\begin{tabular}[c]{@{}l@{}}Standard \\Learning\end{tabular}}                                     & \multicolumn{2}{l||}{\begin{tabular}[c]{@{}l@{}}Transfer\\Learning\end{tabular}}                                    & \multicolumn{2}{l|}{\begin{tabular}[c]{@{}l@{}}Standard \\Learning\end{tabular}}                                      & \multicolumn{2}{l||}{\begin{tabular}[c]{@{}l@{}}Transfer \\Learning\end{tabular}}                                     & \multicolumn{2}{l|}{\begin{tabular}[c]{@{}l@{}}Standard \\Learning\end{tabular}}                                    & \multicolumn{2}{l|}{\begin{tabular}[c]{@{}l@{}}Transfer\\Learning\end{tabular}}                                        \\ 
\hline
Epoch                  & \begin{tabular}[c]{@{}l@{}}Avg.\\Acc.\\(\%)\end{tabular} & \begin{tabular}[c]{@{}l@{}}Avg. \\Time \\(S)\end{tabular} & \begin{tabular}[c]{@{}l@{}}Avg. \\Acc.\\(\%)\end{tabular} & \begin{tabular}[c]{@{}l@{}}Avg. \\Time \\(S)\end{tabular} & \begin{tabular}[c]{@{}l@{}}Avg.\\Acc. \\(\%)\end{tabular} & \begin{tabular}[c]{@{}l@{}}Avg. \\Time\\(S)\end{tabular} & \begin{tabular}[c]{@{}l@{}}Avg.\\Acc.\\~(\%)\end{tabular} & \begin{tabular}[c]{@{}l@{}}Avg.\\Time\\(S)\end{tabular} & \begin{tabular}[c]{@{}l@{}}Avg.~\\Acc.\\(\%)\end{tabular} & \begin{tabular}[c]{@{}l@{}}Avg. \\Time\\~(S)\end{tabular} & \begin{tabular}[c]{@{}l@{}}Avg. \\Acc.\\(\%)\end{tabular} & \begin{tabular}[c]{@{}l@{}}Avg. \\Time\\~(S)\end{tabular} & \begin{tabular}[c]{@{}l@{}}Avg.\\Acc.\\(\%)\end{tabular} & \begin{tabular}[c]{@{}l@{}}Avg.\\Time\\~(S)\end{tabular} & \begin{tabular}[c]{@{}l@{}}Avg. \\Acc.\\~(\%)\end{tabular} & \begin{tabular}[c]{@{}l@{}}Avg.\\Time\\~(S)\end{tabular}  \\ 
\hline
250                    & 83.50                                                    & 33.15                                                     & 84.08                                                     & 27.03                                                     & 70.94                                                     & 33.44                                                    & 76.77                                                     & 27.44                                                   & 73.54                                                     & 37.99                                                     & 76.24                                                     & 26.94                                                     & 82.68                                                    & 42.81                                                    & 84.43                                                      & 27.75                                                     \\
200                    & 82.72                                                    & 26.48                                                     & 84.26                                                     & 23.11                                                     & 72.88                                                     & 27.38                                                    & 76.97                                                     & 23.39                                                   & 74.32                                                     & 31.40                                                     & 78.43                                                     & 23.41                                                     & 82.95                                                    & 35.76                                                    & 84.13                                                      & 23.65                                                     \\
150                    & 84.19                                                    & 21.28                                                     & 84.36                                                     & 19.27                                                     & 77.24                                                     & 22.10                                                    & 77.00                                                     & 19.48                                                   & 74.59                                                     & 24.59                                                     & 80.25                                                     & 19.13                                                     & 84.17                                                    & 27.53                                                    & 84.39                                                      & 19.46                                                     \\
100                    & 83.37                                                    & 15.24                                                     & 82.41                                                     & 14.97                                                     & 77.25                                                     & 15.60                                                    & 75.60                                                     & 15.09                                                   & 68.89                                                     & 17.74                                                     & 82.91                                                     & 14.84                                                     & 84.38                                                    & 18.94                                                    & 82.13                                                      & 14.97                                                     \\
50                     & 86.42                                                    & 9.52                                                      & 81.06                                                     & 10.68                                                     & 77.47                                                     & 10.01                                                    & 73.17                                                     & 10.87                                                   & 79.42                                                     & 10.90                                                     & 79.39                                                     & 10.71                                                     & 84.58                                                    & 11.60                                                    & 80.91                                                      & 10.72                                                     \\
\hline
\end{tabular}
\end{table*}

%% file: tables/lstm_en_de_atn.tex
\begin{table*}
\centering
\caption{Results summary in target domain using LSTM Encoder Decoder with Attention layer Model}
\label{tab:taget_domain_en_de_atn}
\begin{tabular}{|l|l|l|l|l||l|l|ll||l|l|l|l||l|l|l|l|} 
\hline
\multicolumn{1}{|c|}{\begin{tabular}[c]{@{}c@{}}\\\end{tabular}} & \multicolumn{4}{c||}{Dataset B}                                                                                                                                                                                                              & \multicolumn{4}{c||}{Dataset C}                                                                                                                                                                                                                 & \multicolumn{4}{c||}{Dataset D}                                                                                                                                                                                                                 & \multicolumn{4}{c|}{Dataset E}                                                                                                                                                                                                                 \\ 
\hline
\multicolumn{1}{|c|}{}                                           & \multicolumn{2}{c|}{\begin{tabular}[c]{@{}c@{}}Standard \\Learning\end{tabular}}                                      & \multicolumn{2}{c||}{\begin{tabular}[c]{@{}c@{}}Transfer\\Learning\end{tabular}}                                     & \multicolumn{2}{c|}{\begin{tabular}[c]{@{}c@{}}Standard \\Learning\end{tabular}}                                       & \multicolumn{2}{c||}{\begin{tabular}[c]{@{}c@{}}Transfer \\Learning\end{tabular}}                                      & \multicolumn{2}{c|}{\begin{tabular}[c]{@{}c@{}}Standard \\Learning\end{tabular}}                                      & \multicolumn{2}{c||}{\begin{tabular}[c]{@{}c@{}}Transfer \\Learning\end{tabular}}                                       & \multicolumn{2}{c|}{\begin{tabular}[c]{@{}c@{}}Standard \\Learning\end{tabular}}                                     & \multicolumn{2}{c|}{\begin{tabular}[c]{@{}c@{}}Transfer \\Learning\end{tabular}}                                        \\ 
\hline
Epoch                                                            & \begin{tabular}[c]{@{}l@{}}Avg.\\Acc. \\(\%)\end{tabular} & \begin{tabular}[c]{@{}l@{}}Avg. \\Time \\(S)\end{tabular} & \begin{tabular}[c]{@{}l@{}}Avg.\\Acc.\\(\%)\end{tabular} & \begin{tabular}[c]{@{}l@{}}Avg. \\Time \\(S)\end{tabular} & \begin{tabular}[c]{@{}l@{}}Avg. \\Acc. \\(\%)\end{tabular} & \begin{tabular}[c]{@{}l@{}}Avg. \\Time \\(S)\end{tabular} & \begin{tabular}[c]{@{}l@{}}Avg.\\Acc. \\(\%)\end{tabular} & \begin{tabular}[c]{@{}l@{}}Avg. \\Time\\(S)~~\end{tabular} & \begin{tabular}[c]{@{}l@{}}Avg. \\Acc. \\(\%)\end{tabular} & \begin{tabular}[c]{@{}l@{}}Avg.\\Time \\(S)\end{tabular} & \begin{tabular}[c]{@{}l@{}}Avg. \\Acc.~ \\(\%)\end{tabular} & \begin{tabular}[c]{@{}l@{}}Avg.\\Time\\(S)~~\end{tabular} & \begin{tabular}[c]{@{}l@{}}Avg.\\Acc.\\(\%)\end{tabular} & \begin{tabular}[c]{@{}l@{}}Avg. \\Time \\(S)\end{tabular} & \begin{tabular}[c]{@{}l@{}}Avg. \\Acc. \\(\%)\end{tabular} & \begin{tabular}[c]{@{}l@{}}Avg. \\Time \\(S)\end{tabular}  \\ 
\hline
250                                                              & 83.55                                                     & 22.51                                                     & 83.53                                                    & 23.82                                                     & 74.10                                                      & 25.14                                                     & 75.19                                                     & 25.22                                                      & 71.91                                                      & 28.23                                                    & 77.71                                                       & 24.45                                                     & 82.42                                                    & 28.37                                                     & 83.76                                                      & 24.17                                                      \\
200                                                              & 83.44                                                     & 18.07                                                     & 83.68                                                    & 20.11                                                     & 72.07                                                      & 20.62                                                     & 74.78                                                     & 20.55                                                      & 75.01                                                      & 23.16                                                    & 81.70                                                       & 19.94                                                     & 84.77                                                    & 22.96                                                     & 83.62                                                      & 20.07                                                      \\
150                                                              & 84.98                                                     & 15.06                                                     & 83.77                                                    & 15.70                                                     & 75.42                                                      & 16.16                                                     & 75.79                                                     & 16.54                                                      & 72.61                                                      & 18.43                                                    & 83.57                                                       & 15.95                                                     & 84.36                                                    & 19.12                                                     & 83.49                                                      & 15.70                                                      \\
100                                                              & 83.72                                                     & 10.83                                                     & 80.85                                                    & 11.53                                                     & 76.46                                                      & 11.73                                                     & 74.96                                                     & 12.14                                                      & 70.32                                                      & 13.35                                                    & 84.29                                                       & 11.70                                                     & 85.02                                                    & 13.45                                                     & 80.46                                                      & 11.61                                                      \\
50                                                               & 86.80                                                     & 7.24                                                      & 78.80                                                    & 7.23                                                      & 77.43                                                      & 7.73                                                      & 72.93                                                     & 7.78                                                       & 79.07                                                      & 8.54                                                     & 81.43                                                       & 7.06                                                      & 84.91                                                    & 8.64                                                      & 78.54                                                      & 7.19                                                       \\
\hline
\end{tabular}
\end{table*}

%% file: tables/lstm_en_de_fig.tex

\begin{figure}
    \centering
    \includegraphics[height=4cm,width=9cm]{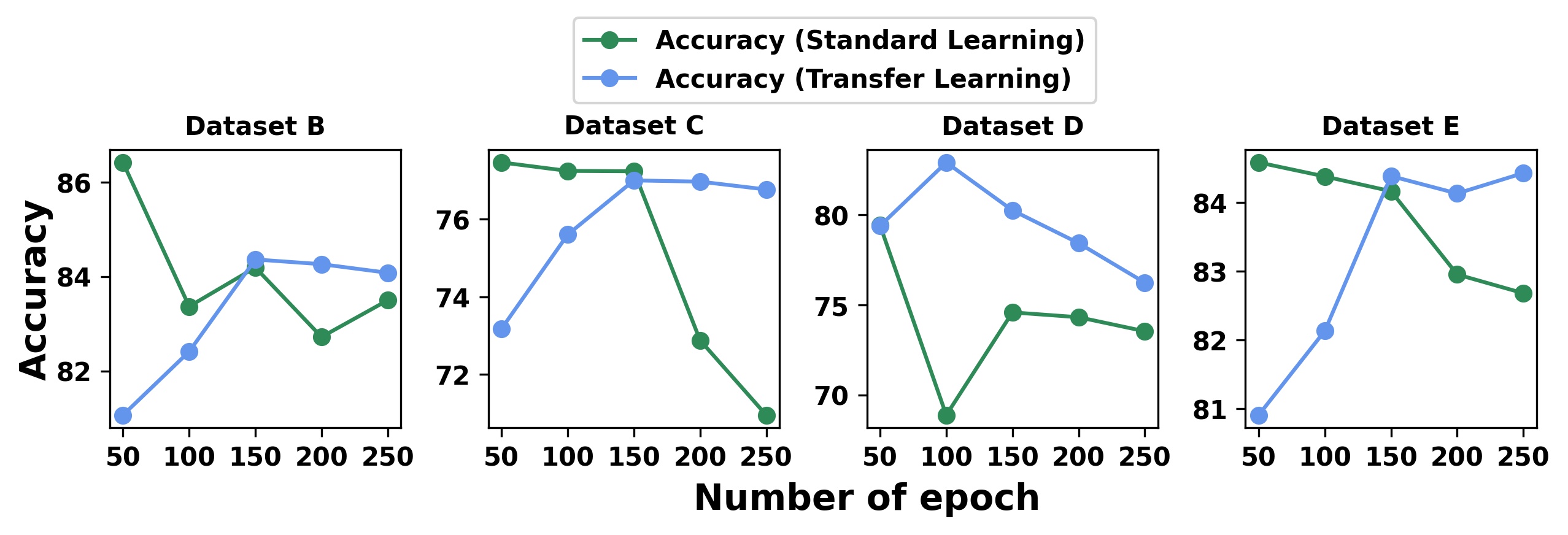}
    \caption{LSTM\_En\_De model accuracy comparison}
    \label{fig:lstm en de model accuracy}
\end{figure}
\begin{figure}
    \centering
    \includegraphics[height=4cm,width=9cm]{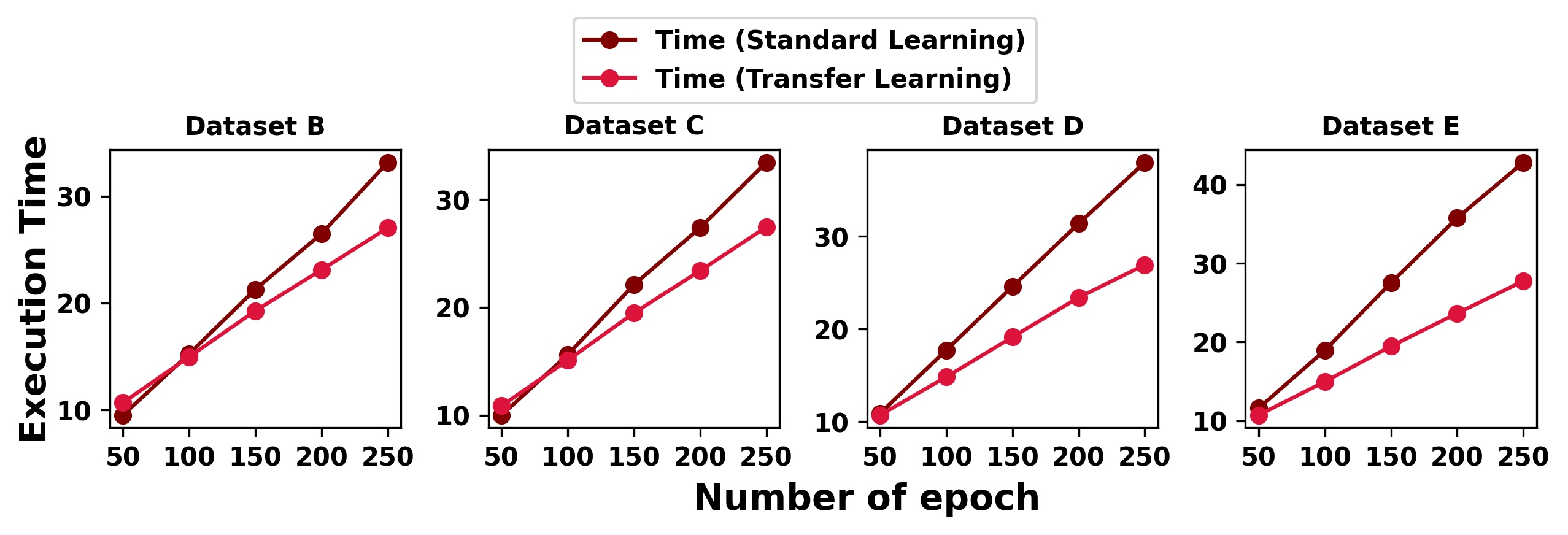}
    \caption{LSTM\_En\_De model execution time comparison}
    \label{fig:lstm en de model time}
\end{figure}

%% file: tables/lstm_en_de_atn_fig.tex

\begin{figure}
    \centering
    \includegraphics[height=4cm,width=9cm]{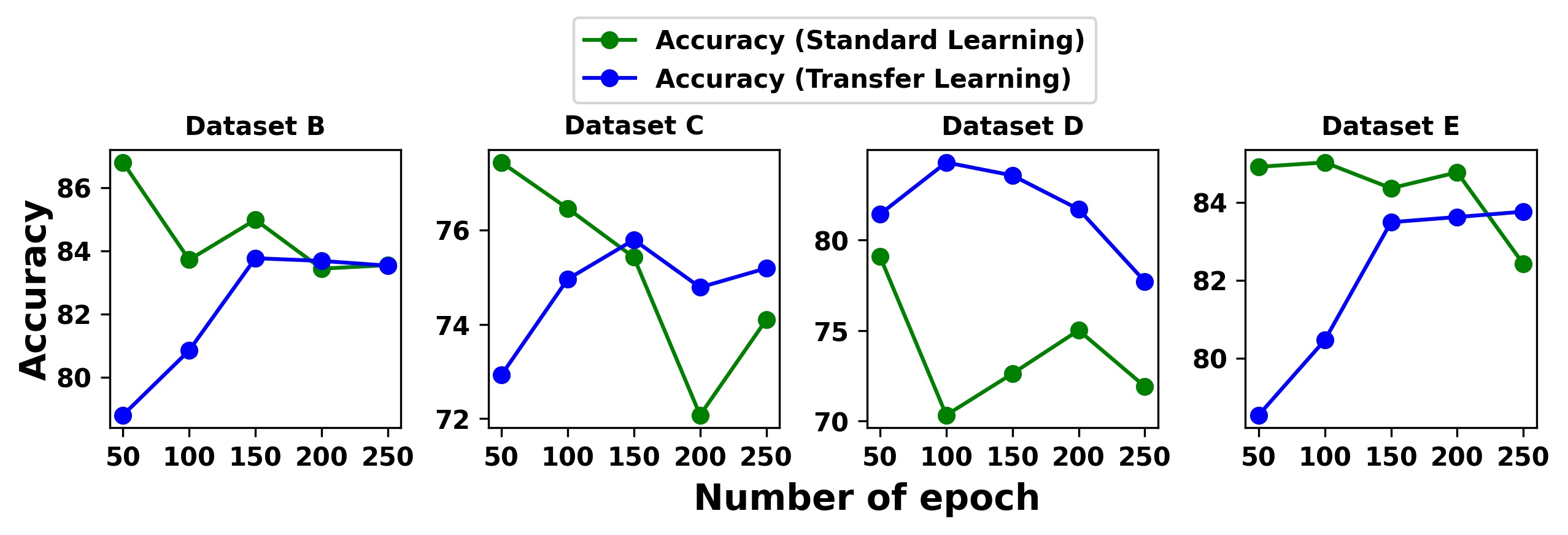}
    \caption{LSTM\_En\_De\_Atn model accuracy comparison}
    \label{fig:lstm en de atn model accuracy}
\end{figure}
\begin{figure}
    \centering
    \includegraphics[height=4cm,width=9cm]{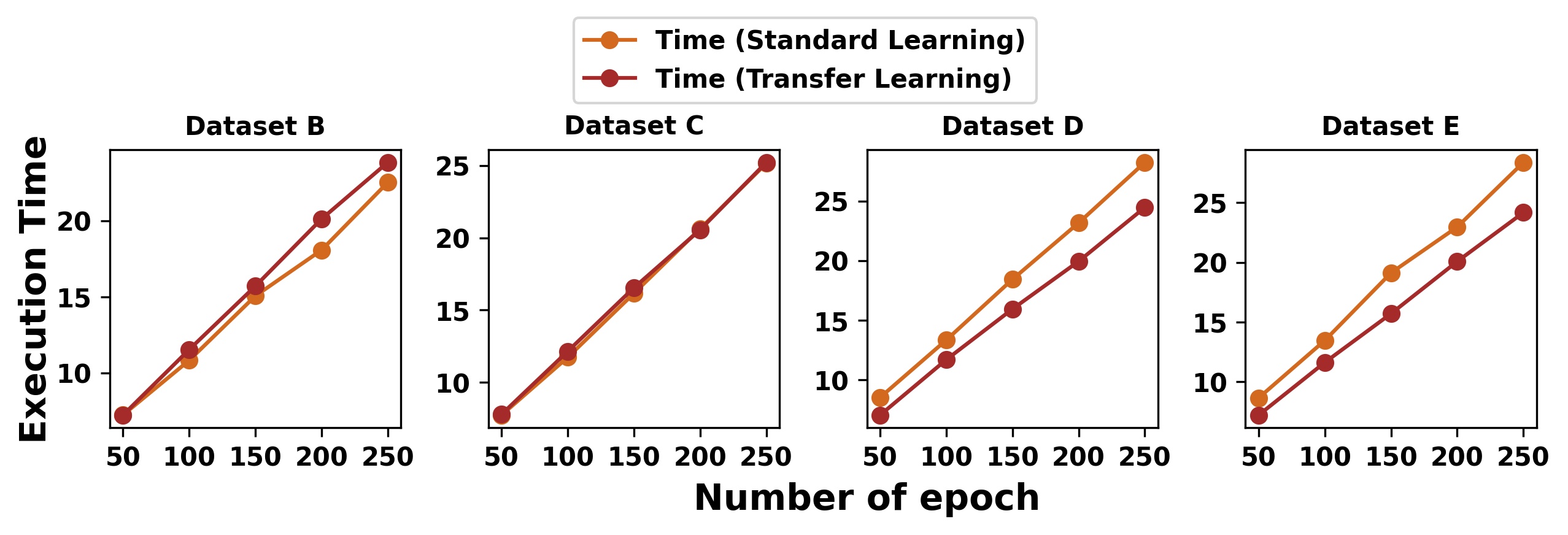}
    \caption{LSTM\_En\_De\_Atn model execution time comparison}
    \label{fig:lstm en de atn model time}
\end{figure}